\documentclass[10pt,twocolumn,letterpaper]{article}

\usepackage{cvpr}
\usepackage{times}
\usepackage{epsfig}
\usepackage{graphicx}
\usepackage{amsmath}
\usepackage{amssymb}
\usepackage{subfig}
\usepackage{authblk}

\usepackage[breaklinks=true,bookmarks=false]{hyperref}

\cvprfinalcopy % *** Uncomment this line for the final submission

\def\cvprPaperID{213} % *** Enter the CVPR Paper ID here

\newcommand{\norm}[1]{\left\lVert#1\right\rVert}

\begin{document}
\setlength{\affilsep}{1ex}
%%%%%%%%% TITLE
\title{Attention-Aware Face Hallucination via Deep Reinforcement Learning\vspace{-3ex}}

\author{Qingxing Cao, \quad Liang Lin\thanks{Corresponding author is Liang Lin. This work was supported in part by the State Key Development Program under Grant 2016YFB1001004, in part by the National Natural Science Foundation of China under Grant 61622214. This work was also supported by Special Program for Applied Research on Super Computation of the NSFC-Guangdong Joint Fund (the second phase).}, \quad Yukai Shi, \quad Xiaodan Liang, \quad Guanbin Li\\
	School of Data and Computer Science, Sun Yat-sen University, Guangzhou, China\\
	{\tt\footnotesize caoqx@mail2.sysu.edu.cn}, {\tt\small linliang@ieee.org}, {\tt\small shiyk3@mail2.sysu.edu.cn}, {\tt\small xdliang328@gmail.com}, {\tt\small liguanbin@mail.sysu.edu.cn}
	\vspace{-3ex}
}

% \author{Qingxing Cao$^1$, Liang Lin$^1$, Yukai Shi$^1$, Xiaodan Liang$^2$, Guanbin Li$^1$\\
% 	$^1$Sun Yat-sen University \quad $^2$Carnegie Mellon University\\
% 	{\tt\footnotesize caoqx@mail2.sysu.edu.cn}, {\tt\small linliang@ieee.org}, {\tt\small shiyk3@mail2.sysu.edu.cn}, {\tt\small xiaodan1@cs.cmu.edu}, {\tt\small liguanbin@mail.sysu.edu.cn}
% 	\vspace{-3ex}
% }

%\author[1]{Qingxing Cao}
%\author[1]{Liang Lin}
%\author[1]{Yukai Shi}
%\author[1]{Guanbin Li}
%\author[2]{Xiaodan Liang}
%\affil[1]{Sun Yat-sen University}
%\affil[2]{Carnegie Mellon University}

%\renewcommand\Authfont{\normalsize}
%\renewcommand\Affilfont{\normalsize}
%\renewcommand\Authsep{, }
%\renewcommand\Authand{ and }

%\author{Qingxing Cao\\
%Sun Yat-sen University\\
%{\tt\small caoqx@mail2.sysu.edu.cn}
%\and
%Liang Lin\\
%Sun Yat-sen University\\
%{\tt\small linliang@ieee.org}
%\and
%Yukai Shi\\
%Sun Yat-sen University\\
%{\tt\small shiyk3@mail2.sysu.edu.cn}
%\and
%Guanbin Li\\
%Sun Yat-sen University\\
%\and
%Xiaodan Liang\\
%Carnegie Mellon University\\
%{\tt\small xiaodan1@cs.cmu.edu}
%	% For a paper whose authors are all at the same institution,
%	% omit the following lines up until the closing ``}''.
%	% Additional authors and addresses can be added with ``\and'',
%	% just like the second author.
%	% To save space, use either the email address or home page, not both
%}

\maketitle
\thispagestyle{empty}

%%%%%%%%% ABSTRACT
\begin{abstract}
Face hallucination is a domain-specific super-resolution problem with the goal to generate high-resolution (HR) faces from low-resolution (LR) input images. In contrast to existing methods that often learn a single patch-to-patch mapping from LR to HR images and are regardless of the contextual interdependency between patches, we propose a novel Attention-aware Face Hallucination (Attention-FH) framework which resorts to deep reinforcement learning for sequentially discovering attended patches and then performing the facial part enhancement by fully exploiting the global interdependency of the image. Specifically, in each time step, the recurrent policy network is proposed to dynamically specify a new attended region by incorporating what happened in the past. The state (i.e., face hallucination result for the whole image) can thus be exploited and updated by the local enhancement network on the selected region. The Attention-FH approach jointly learns the recurrent policy network and local enhancement network through maximizing the long-term reward that reflects the hallucination performance over the whole image. Therefore, our proposed Attention-FH is capable of adaptively personalizing an optimal searching path for each face image according to its own characteristic. Extensive experiments show our approach significantly surpasses the state-of-the-arts on in-the-wild faces with large pose and illumination variations.
\end{abstract}

%%%%%%%%% BODY TEXT
\section{Introduction}

Face hallucination refers to generating a high-resolution face image from a low-resolution input image, which is a very fundamental problem in face analysis field and can facilitate several face-related tasks such as face attribute recognition~\cite{liu2015deep}, face alignment~\cite{zhang2016learning}, face recognition~\cite{zhou2015naive} in the complex real-world scenarios in which the face images are often of very low quality.

Existing face hallucination methods~\cite{yang2010image,yang2013structured,ma2010hallucinating,tappen2012bayesian} usually focus on how to learn a discriminative patch-to-patch mapping from LR images to HR images. Particularly, recent great progresses are made by employing the advanced Convolutional Neural Networks (CNNs)~\cite{zhou2015learning} and multiple cascaded CNNs~\cite{zhu2016deep}. The face structure priors and spatial configurations~\cite{Liu2007,dong2014learning} are often treated as external information for enhancing faces / facial parts. However, the contextual dependencies among the facial parts are usually ignored during the hallucination processing. According to the studies of human perception process~\cite{najemnik2005optimal}, humans start with perceiving the whole images and successively explore a sequence of regions with the attention shifting mechanism, rather than separately processing the local regions. This finding enlightens us to explore a new pipeline of face hallucination by sequentially searching for the attentional local regions and considering their contextual dependency from a global perspective.

\begin{figure*}[t]
	\centering
	\includegraphics[width=1.0 \textwidth]{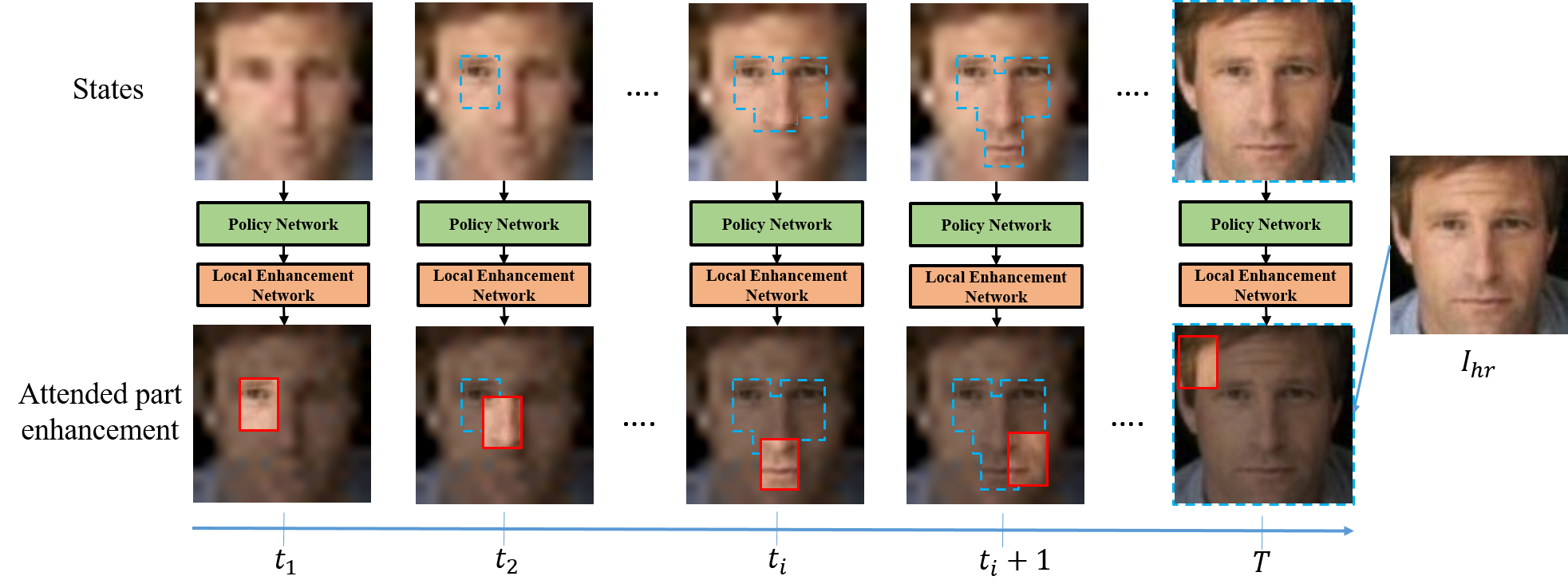}
	
	\caption{Sequentially discovering and enhancing facial parts in our Attention-FH framework. At each time step, our framework specifies an attended region based on the past hallucination results and enhances it by considering the global perspective of the whole face. The red solid bounding boxes indicate the latest perceived patch in each step and the blue dashed bounding boxes indicate all the previously enhanced regions. We adopt a global reward at the end of sequence to drive the framework learning under Reinforcement Learning paradigm.}
	\label{fig:intro}
	\vspace{-3mm}
\end{figure*}

Inspired by the recent successes of attention and recurrent models on a variety of computer vision tasks~\cite{sun2015deepid3,caicedo2015active,DRAW}, we propose an Attention-aware face hallucination (Attention-FH) framework that recurrently discovers facial parts and enhance them by fully exploiting the global interdependency of the image, as shown in Fig.~\ref{fig:intro}. In particular, accounting for the diverse characteristics of face images on blurriness, pose, illumination and face appearance, we explore to search for an optimal accommodated enhancement route for each face hallucination. And we resort to the deep reinforcement learning (RL) method ~\cite{alphaGo} to harness the model learning since this technique has been demonstrated its effectiveness on globally optimizing the sequential models without supervision for every step.

Specifically, our Attention-FH framework jointly optimizes a recurrent policy network that learns the policies to select a preferable facial part in each step and a local enhancement network for facial parts hallucination, through considering the previous enhancement results of the whole face. In this way, rich correlation cues among different facial parts can be explicitly incorporated in the local enhancement process in each step. For example, the agent can improve the enhancement of the mouth region by taking a more clear version of the eye region into account, as Fig.~\ref{fig:intro} illustrates. 

%Benefiting from the joint learning of recurrent policy network and local enhancement network, our framework can take advantage both the representative power of CNNs in a local region, and long-term memorization capability of LSTM~\cite{LSTM} to enable the long-range inference.

We define the global reward for reinforcement learning by the overall performance of the super-resolved face, which drives the recurrent policy network optimization. And the recurrent policy network is optimized following the reinforcement learning (RL) procedure~\cite{REINFORCE} that can be treated as a Markov decision process (MDP) maximized with a long-term global reward. In each time step, we learn the policies to determine an optimal location of next attended region by conditioning on the current enhanced whole face and the history actions. One Long Short-Term Memory (LSTM) layer is utilized for capturing the past information of the attended facial parts. And the history actions are also memorized to avoid the inference trapped in a repetitive action cycle.

Given the selected facial part in each step, the local enhancement network performs the super-resolution operation. The loss of enhancement is defined based on the facial part hallucination quality. Notably, the supervision information from  the enhancement of facial parts effectively reduces unnecessary trials and errors during the reinforcement optimization.

We compare the proposed Attention-FH approach with other state-of-the-art face hallucination methods under both constrained and unconstrained settings. Extensive experiments have shown that our method substantially surpasses all of them. Moreover, our framework can explicitly generate a sequence of attentional regions during the hallucination, which finely accord with human perception process.

\section{Related Work}
\textbf{Face Hallucination and Image Super-Resolution.} Face hallucination problem is a special case of image super-resolution, which requires more informative structure priors and suffers from more challenging blurring. Early techniques made an assumption that the faces are in a controlled setting with small variations. Wang \emph{et~al.}~\cite{wang2005hallucinating} implemented the mapping between low-resolution and high-resolution faces by an eigen transformation. Yang \emph{et~al.}~\cite{yang2010image} suggested that the low-resolution and high-resolution faces have similar sparse prior and the high-resolution faces can be accurately recovered from the low-dimensional projections. In particular, Yang \emph{et~al.}~\cite{yang2013structured} replaced the patch-to-patch mapping with mapping between specific facial components, which incorporates priors on face. However, the matchings between components are based on the landmark detection results which are unavailable when the down-sampling factor is large. Recently, deep convolution neural network has been successfully applied to face hallucination~\cite{zhu2016deep,zhou2015learning} as well as image super-resolution~\cite{deepSR,ren2015shepard}. Zhen \emph{et~al.}~\cite{cui2014deep} advocated the use of network cascade for image SR with a local auto-encoder architecture. Similar FCNs were also used in~\cite{dong2014learning}, which formulated sparse-coding based SR method into 3 layers convolution neural network. Zhou \emph{et~al.}~\cite{zhou2015learning} addressed the importance of appearance invariant method, and adopted fully-connected layer to perform face hallucination. Ren  \emph{et~al.}~\cite{ren2015shepard} suggested that the unevenly distributed pixels may have different influences. They used Shepard interpolation to efficiently achieve translation invariant interpolation(TVI).

%\subsection{Recurrent generative model}
%The recurrent model has been recently proved to be effective
%\cite{recurrentSR} \cite{recurrentSaliency}

\textbf{Reinforcement Learning and Attention Networks.} Attention mechanism has been recently applied and has benefited various tasks, such as object proposal\cite{NIPS2016_6532}, object classification\cite{NIPS_attention}, relationship detection\cite{relDet}, image captioning\cite{captionAttn} and visual question answering\cite{vqa}. Since contextual information is important for computer vision problems, most of these works attempted to attend multiple regions by formulating their attention procedure as a sequential decision problem. Reinforcement learning technique was introduced to optimize the sequential model with delayed reward. This technique has been applied to face detection\cite{FaceDetRL} and object localization\cite{caicedo2015active}. These methods learned an agent that actively locates the target regions (objects) instead of exhaustively sliding sub-windows on images. For example, Goodrich \emph{et~al.}~\cite{FaceDetRL} defined 32 actions to shift the focal point and reward the agent when finding the goal. Caicedo \emph{et~al.}~\cite{caicedo2015active} defined an action set that contains several transformations of the bounding box and rewarded the agent if the bounding box is closer to the ground-truth in each step. These two methods both learned an optimal policy to locate the target through Q-learning.

\section{Attention-Aware Face Hallucination}
\begin{figure*}[t]
	\centering
	\includegraphics[width=0.90\textwidth]{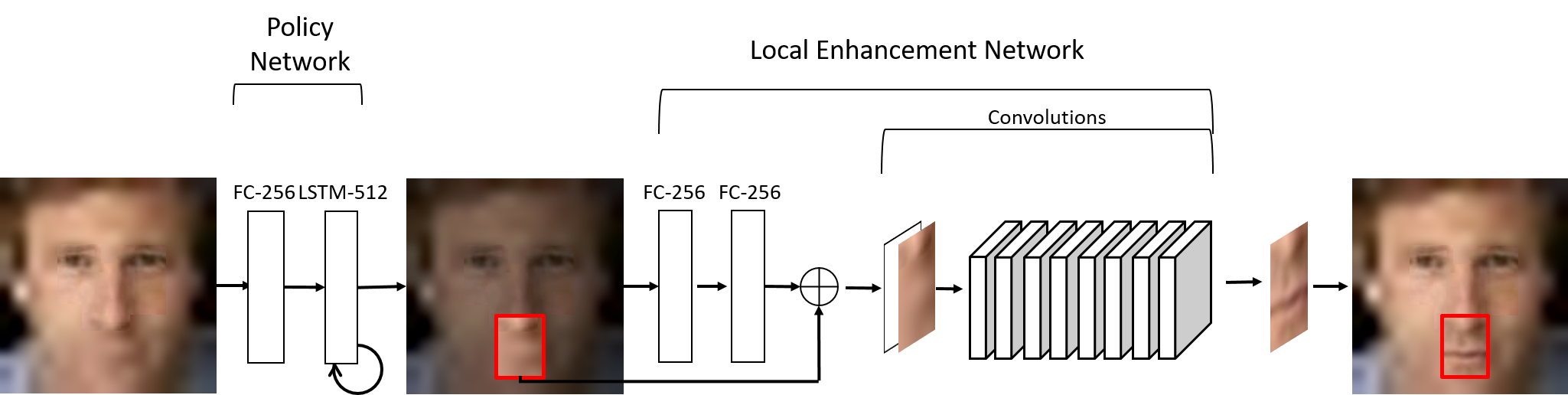}
	\caption{Network architecture of our recurrent policy network and local enhancement network. At each time step, the recurrent policy network takes a current hallucination result $I_{t-1}$ and action history vector encoded by LSTM (512 hidden states) as the input and then outputs the action probabilities for all $W\times H$ locations, where W, H are the height and width of the input image. The policy network first encodes the $I_{t-1}$ with one fully-connected layer (256 neurons), and then fuse the encoded image and the action vector with a LSTM layer. Finally a fully-connected linear layer is appended to generate the $W\times H$-way probabilities. Given the probability map, we extract the local patch, then pass the patch and $I_{t-1}$ into the local enhancement network to generate the enhanced patch. The local enhancement network is constructed by two fully-connected layers (each with 256 neurons) for encoding $I_{t-1}$ and $8$ cascaded convolutional layers for image patch enhancement. Thus a new face hallucination result can be generated by replacing the local patch with an enhanced patch.}
	\label{fig:model_overview}
	\vspace{-3mm}
\end{figure*}

Given a face image $I_{lr}$ with low-resolution, our Attention-FH framework targets on its corresponding high-resolution face image $I_{hr}$ by learning a projection function $F$:
\begin{equation}
I_{hr} = F(I_{lr}|\theta),
\end{equation}
where $\theta$ denotes the function parameters. Our Attention-FH proposes to sequentially locate and enhance the attended facial parts in each step, which can be formulated as a deep reinforcement learning procedure. Our framework consists of two networks: the recurrent policy network that dynamically determines the specific facial part to be enhanced in current step and the local enhancement network which is further employed to enhance the selected facial part. 

Specifically, the whole hallucination procedure of our Attention-FH can be formulated as follows. Given the input image $I_{t-1}$ at the $t$-th step, the agent of the recurrent policy network selects one local facial part $\hat{I}_{t-1}^{l_t}$ with the location $l_t$:
\begin{equation}
\begin{aligned}
l_t &= f_\pi(s_{t-1};\theta_\pi),  \\
\hat{I}_{t-1}^{l_t} &= g(l_t,I_{t-1}),
\end{aligned}
\end{equation}
where $f_\pi$ represents the recurrent policy network, $\theta_\pi$ is the network parameters. $s_{t-1}$ is the encoded input state of the recurrent policy network, which is constructed by the input image $I_{t-1}$ and the encoded history action $h_{t-1}$. $g$ denotes a cropping operation which crops a fixed-size patch from $I_{t-1}$ at location $l_t$ as the selected facial part. The patch size is set as $60\times45$ for all face images.

We then enhance each local facial part $\hat{I}_{t-1}^{l_t}$ by our local enhancement network $f_{e}$. The resulting enhanced local patch $\hat{I}_{t}^{l_t}$ is computed as:
\begin{equation}
\hat{I}_{t}^{l_t} = f_{e}(\hat{I}_{t-1}^{l_t},I_{t-1};\theta_e),
\end{equation}
where $\theta_e$ is the local enhancement network parameters. The output image $I_{t}$ at each $t$-th step is thus obtained by replacing the local patch of the input image $I_{t-1}$ at location $l_t$ with the enhanced patch $\hat{I}_{t}^{l_t}$. Our whole sequential Attention-FH procedure can be written as:
\begin{equation}
\begin{cases}
I_0 = I_{lr}               &\\
I_t = f(I_{t-1};\theta)  & 1 \leq t \leq T, \\
I_{hr} = I_T
\end{cases}
\end{equation}
where $T$ is the maximal number of local patch mining steps, $\theta = [\theta_\pi;\theta_e]$ and $f = [f_\pi;f_e]$. We set $T=25$ empirically throughout this paper.

\subsection{Recurrent Policy Network}
The recurrent policy network performs the sequential local patch mining, which can be treated as a decision making process on discrete time intervals. At each time step, the agent takes action to determine an optimal image patch to be enhanced by conditioning on the current state it has reached so far. Given the selected location, the extracted local patch is enhanced through the proposed local enhancement network. During each time step, the state is updated by rendering the hallucinated face image with the enhanced facial part. The policy network recurrently selects and enhances local patches until the maximum time step is achieved. At the end of this sequence, a delayed global reward, which is measured by the mean squared error between the final face hallucination result and groundtruth high-resolution image, is employed to guide the policy learning of the agent. The agent can thus iterate to explore an optimal searching route for each individual face image in order to maximize the global holistic reward.

\textbf{State:} The state $s_{t}$ at $t$-th step should be able to provide enough information for the agent to decide without looking back more than one step. It is thus composed of two parts: 1) the enhanced hallucinated face image $I_{t}$ from previous steps, which enables the agent to sense rich contextual information for a new patch to be processed, e.g., the part which is still blur and requires to be enhanced; 2) the latent variable $h_{t}$ obtained by forwarding the encoded history action vector $h_{t-1}$ into the LSTM layer, which is used to incorporate all previous actions. In this way, the goal of the agent is to determine the location of the next attended local patch by sequentially observing state $s_{t} = \{I_t,h_t\}$ to generate a high-resolution image $I_{hr}$.

\textbf{Action:}
Given a face image $I$ with size $W\times H$, the agent targets on selecting one action from all possible locations $l_t = (x,y|1 \leq x \leq W,1\leq y \leq H)$. As shown in Fig.~\ref{fig:model_overview}, at each time step $t$, the policy network $f_{\pi}$ first encodes the current hallucinated face image $I_{t-1}$ with fully-connected layer. Then the LSTM unit in policy network fuses the encoded vector with the history action vector $h_{t-1}$. Ultimately, a final linear layer is appended to produce a $W\times H$-way vector which indicates the probabilities of all available actions $P(l_{t}=(x,y)|s_{t-1})$, with each entry $(x,y)$ indicating the probability of next attached patch located in position $(x,y)$. The agent will then take action $l_t$ by stochastically drawing an entry following the action probability distribution. During testing, we select the location $l_t$ with the highest probability.

\textbf{Reward:} The reward is applied to guide the agent to learn the sequential policies to obtain the whole action sequence. Since our model targets on generating a hallucinated face image, we define the reward according to  mean squared error (MSE) after enhancing $T$ attended local patches at the selected locations with the local enhancement network. Given the fixed local enhancement network $f_{e}$, we first compute the final face hallucination result $I_T$ by sequentially enhancing a list of local patches mined by $\mathbf{l} = l_{1,2,...,T}$. The MSE loss is thus obtained by computing $L_{\theta_\pi} = E_{p(\mathbf{l};\pi)}[ \norm{I_{hr} - I_T}_2]$, where $p(\mathbf{l};\pi)$ is the probability distribution produced by the policy network $f_{\pi}$ . The reward $r$ at $t$-th step can be set as:

\begin{equation}
r_t = \begin{cases}
0     &t < T\\
-L_{\theta_\pi}  &t = T.
\end{cases}
\end{equation}

By setting the discounted factor as 1, the total discounted reward will be $R = -L_{\theta_\pi}$.

\subsection{Local Enhancement Network}
The local enhancement network $f_e$ is used to enhance the extracted low-resolution patches. Its input contains the whole face image $I_{t-1}$ that is rendered by all previous enhanced results and the selected local patch $\hat{I}_{t-1}^{l_t}$ at current step. As shown in Fig.~\ref{fig:model_overview}, we pass the input $I_{t-1}$ into two fully-connected layers to generate a feature map that has the same size of the extracted patch $\hat{I}_{t-1}^{l_t}$ in order to encode the holistic information of $I_{t-1}$. This feature map is then concatenated with the extracted patch $\hat{I}_{t-1}^{l_t}$ and go through convolution layers to obtain the enhanced patch $\hat{I}_{t}^{l_t}$.

We employ the cascaded convolution network architecture similar to general image super-resolution methods~\cite{dong2014learning, deepSR}. No pooling layers are used between convolution layers, and the sizes of feature maps are kept fixed throughout all convolution layers. 
We follow the detailed setting of the network employed by Tuzel \emph{et~al.}\cite{GLN}. Two fully-connected layers contain 256 neurons. The cascaded convolution network is composed of eight layers. Conv1 and conv7 layers have 16 channels of $3\times 3$ kernels; conv2 and conv6 layers have 32 channels of $7\times 7$ kernels; conv3, conv4 and conv5 layers have 64 channels of size $7\times 7$ kernel; conv8 has kernel of size $5\times 5$ and outputs the enhanced image patch with the same size and channel as the extracted patch.

In the initialization, we first up-sample the image $I_{lr}$ to the same size as high-resolution image $I_{hr}$ with Bicubic method. Our network first generates a residual map and then combines the input low-resolution patch with the residual map to produce the final high-resolution patch. Learning from the residual map has been verified to be more effective than directly learning from the original high-resolution images~\cite{deepSR}~\cite{csc_sr}.

\subsection{Deep Reinforcement Learning}
Our Attention-FH framework jointly trains the parameters $\theta_{\pi}$ of the recurrent policy network $f_{\pi}$ and parameters $\theta_{e}$ of the local enhancement network $f_{e}$. We introduce a reinforcement learning scheme to perform joint optimization.

First, we optimize the recurrent policy network with REINFORCE algorithm~\cite{REINFORCE} guided by the reward given at the end of sequential enhancement. The local enhancement network is optimized with mean squared error between the enhanced patch and the corresponding patch from the ground truth high-resolution image. This supervised loss is calculated at each time step, and can be minimized based on back-propagation.

Since we jointly train the recurrent policy network and local enhancement network, the change of parameters in local enhancement network will affect the final face hallucination result, which in turn causes a non-stationary objective for the recurrent policy network. We further employ the variance reduction strategy as mentioned in~\cite{NIPS_attention} to reduce variance due to the moving rewards during the training procedure.

\section{Experiments}
%To analyze the performance of our recurrent attention-memorized enhancement model, we compared our method with other state-of-the-art methods include generic image super-resolution and face hallucination approaches.
\subsection{Datasets and implementation details} Extensive experiments are evaluated on BioID~\cite{bioid} and LFW~\cite{lfw} datasets. The BioID dataset contains 1521 face images collected under the lab-constrained settings. We use 1028 images for training and 493 images for evaluation.
The LFW dataset contains 5749 identities and 13233 face images taken in an unconstrained environment, in which 9526 images are used for training and the remaining 3707 images are used for evaluation. This train/test split follows the split provided by the LFW datasets. In our experiment, we first align the images on BioID dataset with SDM method~\cite{xiong2013supervised} and then crop the center image patches with size of $160\times120$ as the face images to be processed. For LFW dataset, we use aligned face images provided in LFW funneled\cite{lfw-funneled} and extract the centric $128\times128$ image patches for processing. % are extracted from these images as processed face images.
We evaluate two scaling factors of $4$ and $8$, denoted as 4$\times$ and 8$\times$ in the following figures and tables. The input low resolution image is generated by resizing the original image with fixed scaling factors. Thus the input images in BioID are resized as $40\times30$ and $20\times15$, and the input images in LFW are resized as $32\times32$ and $16\times16$ respectively.

We set the maximum time steps $T = 25$ in our Attention-FH model for both datasets. And the face patch size is $H\times W = 60\times45$ for all experiments. The network is updated using ADAM gradient descent~\cite{adam}. The learning rate and the momentum term are set to 0.0002 and 0.5 respectively.
\begin{figure*}[htbp]\centering
	
	\includegraphics[width=0.90\textwidth]{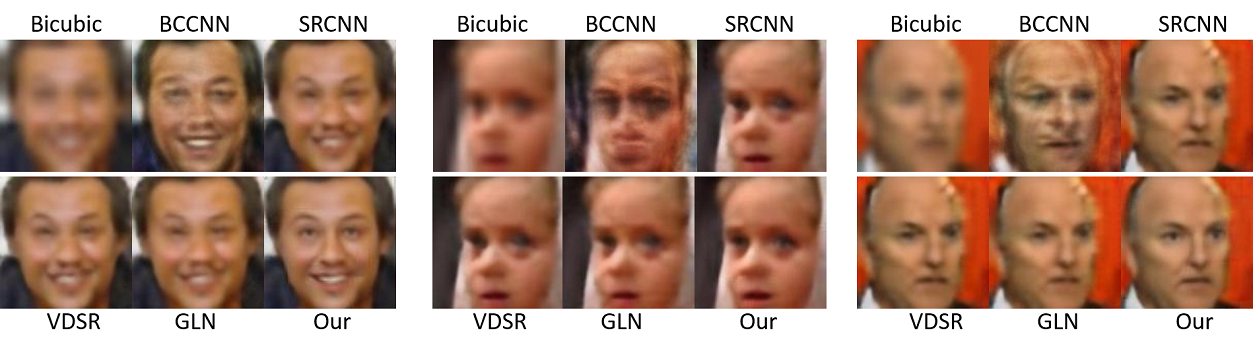}
	
	\includegraphics[width=0.90\textwidth]{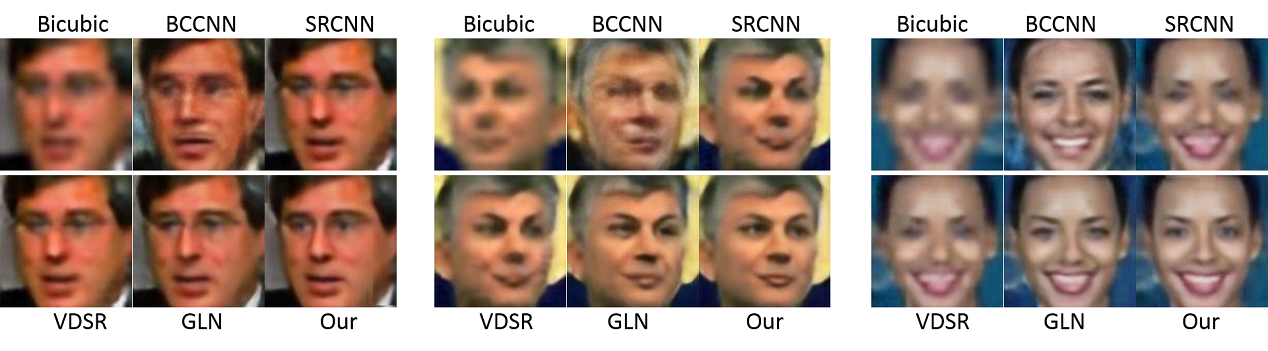}
	\caption{Qualitative results on LFW-funneled with scaling factor of $8$. Results of SFH and MZQ methods are not displayed as they depends on facial landmarks which are often failed to detect in such low-resolution images.}
	\label{fig:visualize_results8x}
	
	\subfloat[][Bicubic]  {
		\includegraphics[width=0.093\textwidth]{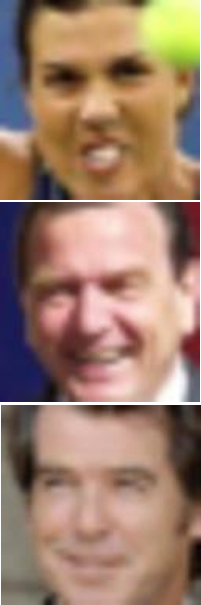}
	}
	\subfloat[][BCCNN]  {
		\includegraphics[width=0.093\textwidth]{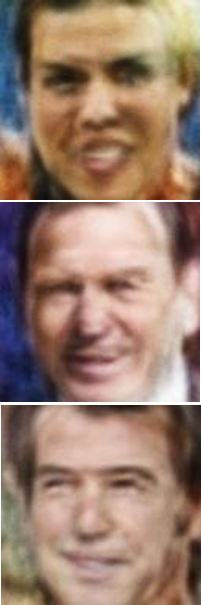}
	}
	\subfloat[][SFH]  {
		\includegraphics[width=0.093\textwidth]{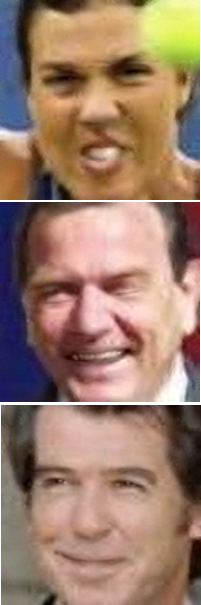}
	}
	\subfloat[][MZQ]  {
		\includegraphics[width=0.093\textwidth]{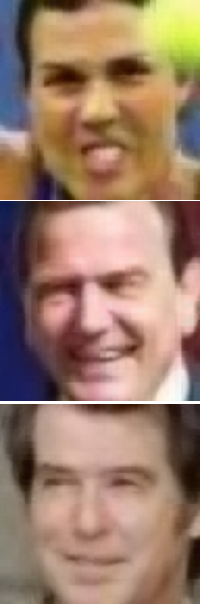}
	}
	\subfloat[][SRCNN]  {
		\includegraphics[width=0.093\textwidth]{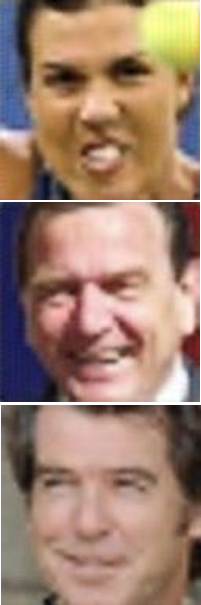}
	}
	\subfloat[][VDSR]  {
		\includegraphics[width=0.093\textwidth]{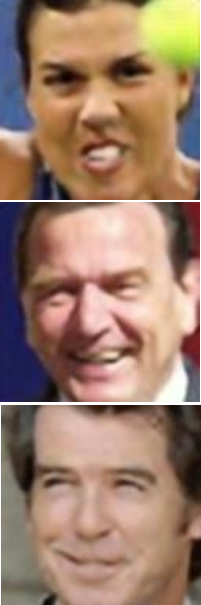}
	}
	\subfloat[][GLN]  {
		\includegraphics[width=0.093\textwidth]{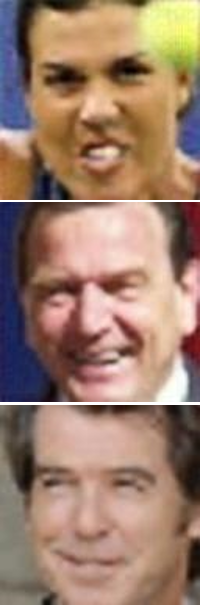}
	}
	\subfloat[][Our]  {
		\includegraphics[width=0.093\textwidth]{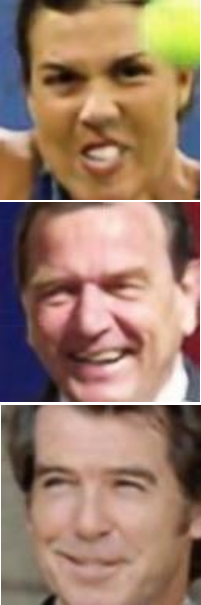}
	}
	\subfloat[][Original]  {
		\includegraphics[width=0.093\textwidth]{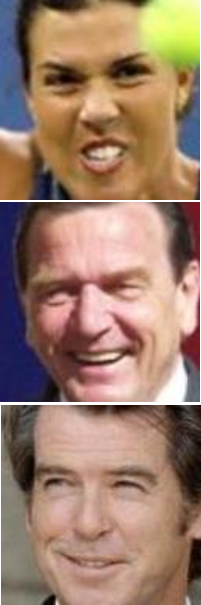}
	}

	\subfloat[][Bicubic]  {
		\includegraphics[width=0.093\textwidth]{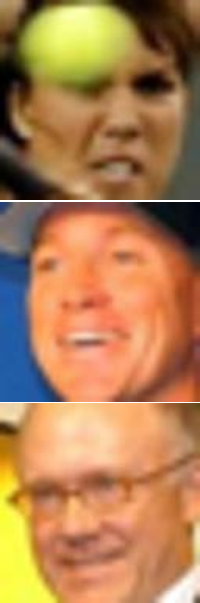}
	}
	\subfloat[][BCCNN]  {
		\includegraphics[width=0.093\textwidth]{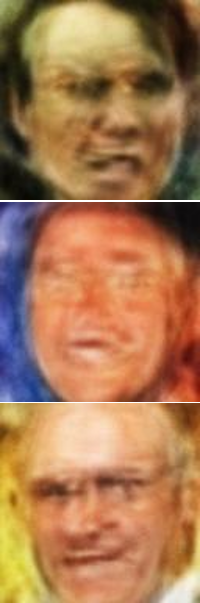}
	}
	\subfloat[][SFH]  {
		\includegraphics[width=0.093\textwidth]{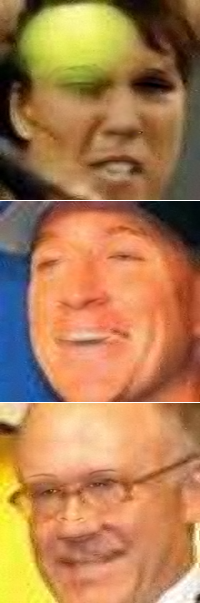}
	}
	\subfloat[][MZQ]  {
		\includegraphics[width=0.093\textwidth]{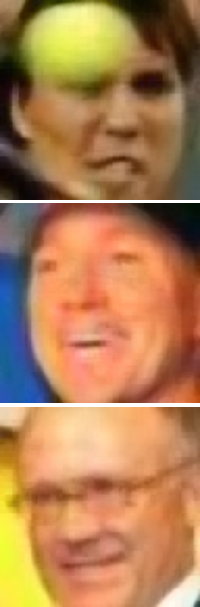}
	}
	\subfloat[][SRCNN]  {
		\includegraphics[width=0.093\textwidth]{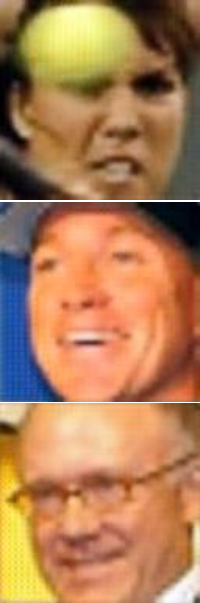}
	}
	\subfloat[][VDSR]  {
		\includegraphics[width=0.093\textwidth]{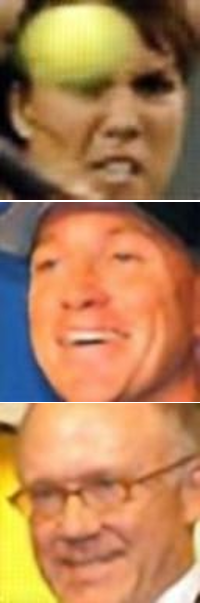}
	}
	\subfloat[][GLN]  {
		\includegraphics[width=0.093\textwidth]{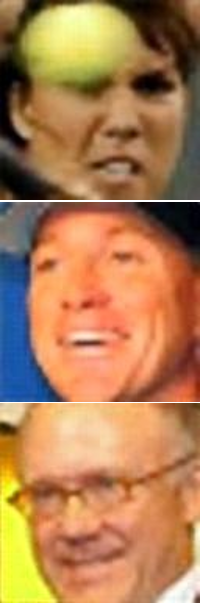}
	}
	\subfloat[][Our]  {
		\includegraphics[width=0.093\textwidth]{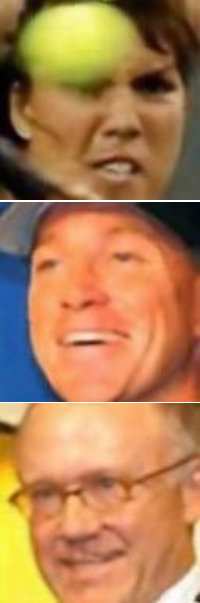}
	}
	\subfloat[][Original]  {
		\includegraphics[width=0.093\textwidth]{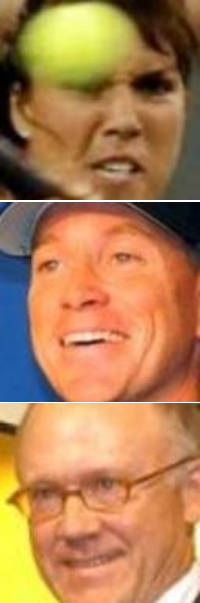}
	}
	\caption{Qualitative results on LFW-funneled with scaling factor of $4$.}
	\label{fig:visualize_results}
	\vspace{-3mm}
	%\vspace{-10pt}
\end{figure*}

\begin{table*}[hbt] \centering \small%\footnotesize %
	\center
	\begin{tabular}{|*{13}{c|}}
		\hline
		Methods & \multicolumn{3}{c|}{LFW-funneled $4\times$} & \multicolumn{3}{c|}{LFW-funneled $8\times$}  &  \multicolumn{3}{c|}{BioID $4\times$} & \multicolumn{3}{c|}{BioID $8\times$}  \\
		
		\hline
		& PSNR & SSIM & FSIM & PSNR & SSIM & FSIM &  PSNR & SSIM & FSIM & PSNR & SSIM & FSIM\\
		\hline\hline
		Bicubic & 26.79 & 0.8469 & 0.8947 & 21.92 & 0.6712 & 0.7824 & 25.18 & 0.8170 & 0.8608 & 20.68 & 0.6434 &0.7539 \\
		\hline
		SFH~\cite{yang2013structured} & 26.59 & 0.8332 & 0.8917 & 22.12 & 0.6732 & 0.7832 &  25.41  & 0.8034& 0.8494 &  20.31 & 0.6234 & 0.7238 \\
		\hline
		BCCNN~\cite{zhou2015learning} & 26.60 & 0.8329 & 0.8982 & 22.62 & 0.6801 & 0.7903 & 24.77 &  0.8034 & 0.8421 & 21.40  & 0.6504 & 0.7621 \\
		\hline
		MZQ~\cite{ma2010hallucinating} & 25.93 & 0.8313 & 0.8865 & 22.12 & 0.6771 & 0.7802 & 24.66  & 0.8003 & 0.8573 & 21.11 & 0.6481  & 0.7594\\
		\hline
		SRCNN~\cite{dong2014learning} & 28.94 & 0.8686 & 0.9069 & 23.92 & 0.6927 & 0.8314 & 27.02  & 0.8517 & 0.8771 & 22.34 & 0.6980 &  0.8274 \\
		\hline
		VDSR~\cite{deepSR} & 29.25 & 0.8711 & 0.9123 & 24.12 & 0.7031 & 0.8391 & 28.52  & 0.8627 & 0.8914 & 24.31  & 0.7321 & 0.8465 \\
		\hline
		GLN~\cite{GLN} & \underline{30.34} & \underline{0.8922} & \underline{0.9151} & \underline{24.51} & \underline{0.7109} & \underline{0.8405} & \underline{29.13} & \underline{0.8794} & \underline{0.8966} & \underline{24.76}  & \underline{0.7421} & \underline{0.8525}\\
		\hline \hline
		Our & \textbf{32.93} & \textbf{0.9104} & \textbf{0.9427}& \textbf{26.17} & \textbf{0.7604} & \textbf{0.8630}& \textbf{31.56} & \textbf{0.9002}& \textbf{0.9343} & \textbf{26.56} & \textbf{0.7864} & \textbf{0.8748}\\
		\hline
	\end{tabular}
	\caption{Comparison between our method and others in terms of PSNR, SSIM and FSIM evaluate metrics. We employ the \textbf{bold face} to label the first place result and \underline{underline} to label the second in each column.}
	\label{table:all_quan_results}
	\vspace{-3mm}
\end{table*}

\subsection{Evaluation protocols and comparisons} We adopt the widely used Peak Signal-to-Noise Ratio (PSNR), structural similarity(SSIM) as well as feature similarity (FSIM)~\cite{zhang2011fsim} as our evaluation measurements.

We compare our method with several state-of-the-art face hallucination and image super-resolution methods. For face hallucination approaches, we compare with SFH~\cite{yang2013structured}, MZQ~\cite{ma2010hallucinating}, BCCNN~\cite{zhou2015learning}, GLN~\cite{GLN}. We also compare with two general image super-resolution methods: SRCNN~\cite{dong2014learning} and VDSR~\cite{deepSR}. For the results of VDSR and SRCNN, we carefully re-implement their models for both scaling factors $4$ and $8$. We first pre-train the models proposed in VDSR and SRCNN using 7,000 images from PASCAL VOC2012~\cite{pascal-voc-2012} and then finetune them on the training sets of LFW and BioID.

\begin{table}[hbt] \centering \small%\footnotesize %
	\center
	\begin{tabular}{|*{5}{c|}}
		\hline
		& LFW $4\times$ & LFW $8\times$  \\
		\hline\hline
		T = 5 & 28.89 & 23.55  \\
		T = 15 & 31.51 & 25.25 \\
		T = 25 & 32.93 & 26.17 \\
		T = 35 & 32.91 & 26.31 \\
		\hline
	\end{tabular}
	\caption{Comparison of the variants of our Attention-FH using different number of steps for sequentially enhancing facial parts on LFW dataset.}
	\vspace{-1mm}
	\label{table:AblationT}
\end{table}

\begin{table}[hbt] \centering \small%\footnotesize %
	\center
	\begin{tabular}{|*{5}{c|}}
		\hline
		& LFW $4\times$ & LFW $8\times$   \\
		\hline\hline
		CNN-16  & 29.11 & 24.02 \\
		Our w/o attention &32.26 & 25.71  \\
		Random Patch &31.60 & 25.76   \\
		$I_{0}$ &32.10 & 25.92   \\
		Spatial Transform &28.13 & 25.75   \\
		\hline
		Our & \textbf{32.93} & \textbf{26.17}  \\
		\hline
	\end{tabular}
	\caption{Comparison of our model with different architecture settings, including the 16-layers convolution network ``CNN-16", the model ``Our w/o attention" that recurrently enhance the whole image, the model that randomly picks patches, the model with original low-resolution input image $I_{0}$ as input for policy network, the end-to-end trainable spatial transform network, and our Attention-FH.}
	\vspace{-5mm}
	\label{table:Ablation}
\end{table}

\subsection{Quantitative and qualitative comparisons} Table~\ref{table:all_quan_results} shows the performance of our model and comparisons with other state-of-the-art methods. Our model substantially beats all compared methods on LFW and BioID datasets in terms of PSNR, SSIM and FSIM metrics. Specifically, the average gains achieved by our model in terms of PSNR are 2.59dB, 1.66dB, 2.43dB and 1.8dB compared with the second best method. The largest improvement by our method is 2.59dB over the the second best method in LFW on scaling factor of $4$.

The traditional face hallucination methods, SFH~\cite{yang2013structured} and MZQ~\cite{ma2010hallucinating}, are highly dependent on the performance of facial landmarks detection. When the scaling factor is set to 8, the landmarks detection results are not reliable given the very low-resolution input, thus their performances are not as good as that of our model. As for deep-learning based methods, our proposed method outperforms the best general image super-resolution method VDSR~\cite{deepSR} by 3.68dB, 2.05dB, 3.04dB and 2.25dB on different experiments respectively. Furthermore, our model outperforms the second best face hallucination method GLN~\cite{GLN} by a significant margin. Noted that GLN~\cite{GLN} shares similar model architecture with our local enhancement network. Therefore, the highly improved performance achieved by our model confirms the effectiveness of utilizing the attention agent.

%\subsection{Qualitative results}
The qualitative comparisons of face hallucination results on LFW dataset and BioID dataset are shown in Fig.~\ref{fig:visualize_results8x} and Fig.~\ref{fig:visualize_results}. As can be observed from these results, our model produces much clearer images than GLN and VDSR, despite the large variations. For example, in Fig.~\ref{fig:visualize_results8x}, the eyes of man on the leftmost column can only be successfully recovered by our Attention-FH, which demonstrates the effectiveness of our recurrent enhanced model.

\subsection{Ablation studies} 
%We apply ablation studies with different network setting to show the effectiveness of the components of our method. 

We perform extensive ablation studies and demonstrate the effects of several important components in our framework:

\textbf{Effectiveness of increasing recursion depth.} Firstly, we explore the effect of using different recursive steps $T$ for sequentially enhancing facial parts.
We train our model with four different settings $T = 5, 15, 25, 35$. Table~\ref{table:AblationT} shows that the face hallucination performance gradually increases with more attention steps. The PSNR measurement improves dramatically when the number of recursion steps is low, since the extracted patches are unable to cover the whole image. When the number of recursion steps gets greater than $15$, the extracted patches can cover the whole image, the step-wise performance improvement on PSNR becomes minor. This phenomenon becomes more obvious when the number of steps gets close to $25$.

\begin{table}[hbt] \centering \footnotesize %
	\vspace{-1mm}
	\center
	\begin{tabular}{|*{4}{c|}}
		\hline
		& SRCNN(3-layers) & VDSR(20-layers) & OURS(8-layers)   \\
		\hline\hline
		Time & 20ms  & 100ms & 81ms \\
		\hline
	\end{tabular}
	\caption{Computational cost for testing on $128\times128$ images.}
	\vspace{-3mm}
	\label{table:time}
\end{table}

\textbf{Effectiveness of patch-wise enhancement manner.} We further evaluate the effectiveness of our patch-wise enhancement manner. In table~\ref{table:Ablation}, ``CNN-16" indicates the results of a 16-layered fully convolution neural network. By comparing our model with ``CNN-16", there are 3.82 dB and 2.14 dB improvements in terms of PSNR on LFW of factor $4$ and factor $8$. We also conduct another ablation study by recurrently enhancing the whole image at each step without extracting patches, named as ``Ours w/o attention". This model has the same architecture as our model, and the number of recurrent steps is set to 5, which nearly covers the same area of overlapping regions as our full model. From table~\ref{table:Ablation}, we can see that although the recurrent model without attention can achieve promising results, our model with attention still promotes 0.67 dB and 0.46 dB on LFW $4\times$ and LFW $8\times$ respectively, which demonstrates the effectiveness of using attention-aware model and reinforcement learning.

\begin{figure*}[ht]
	\centering
	\includegraphics[width=0.93\textwidth]{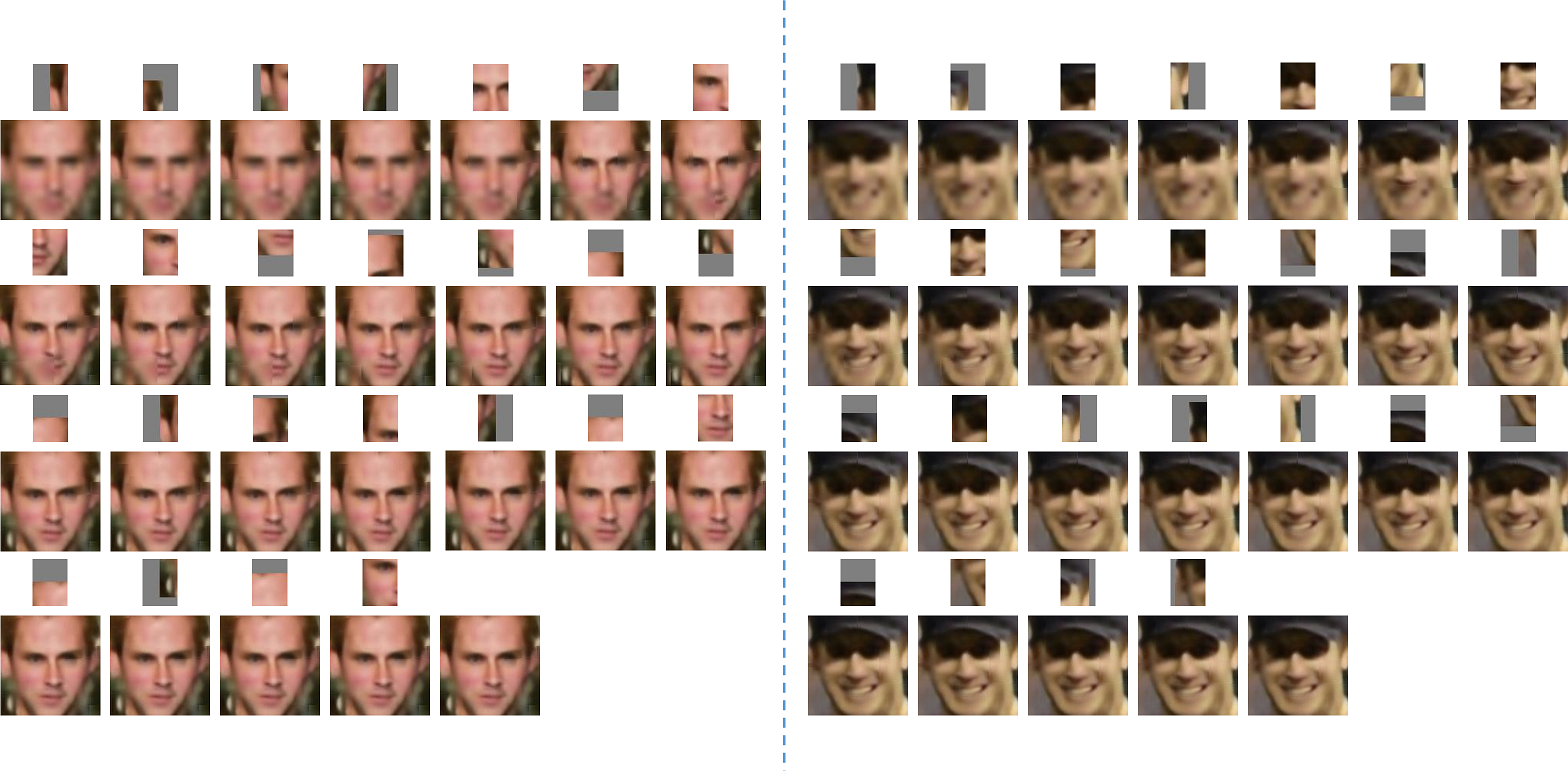}
	
	\caption{Example results of enhancement sequences and corresponding patches selected by the agent. The gray in some patches indicates area outside of original image. Best viewed by zooming in the electronic version.}
	\label{fig:attentive_example}
	\vspace{-4mm}
\end{figure*}

\textbf{Effectiveness of sequentially attending patches.} To demonstrate the ability of our proposed model to sense meaningful attention sequence, we conduct two experiments: 1) The patch is randomly picked at each step instead of being chosen by the agent; 2) use original LR image as input for policy network instead of image that has been locally enhanced at previous steps. As reported in table~\ref{table:Ablation}, randomly picking patches drops 1.33dB and 0.41dB on LFW $4\times$ and $8\times$ respectively, indicating the effectiveness of the agent in our model in locating meaningful attention sequence, which can further be verified in Fig.~\ref{fig:attentive_example}. The second experiment shows that the performance will respectively drop 0.83dB and 0.25dB without previous enhanced information, which confirms that the contextual interdependency between patches can help to mine the next patches by our attention agent.

\textbf{Effectiveness of reinforcement learning.} We also compare our reinforcement learning with end-to-end back-propagation method. Spatial transform network proposed by Jaderberg \emph{et~al.}\cite{st_network} is capable of computing the sub-gradients corresponding to the location of the extracted patch. The comparison results are given in table~\ref{table:Ablation}.

Specifically, for ``spatial transform" model, we replace the softmax output of the policy network with a 2-dimensional vector, which contains coordinate offsets $x$ and $y$ of the extracted patch. The transform scale $s$ is fixed so that the size of extracted patch is also $60\times45$. The spatial transform layer takes $\{x, y, s\}$ as well as the face image generated at previous step as inputs, and extracts the patch for enhancement. We expand the enhanced patch with padding $0$s by the spatial transform layer with a transform vector $\{-s*x, -s*y, 1/s\}$. Then, the expanded patch can be added to the low-resolution face image. We calculate the MSE loss between the groundtruth and the current enhanced result at each step and train the whole recurrent model in an end-to-end manner. Except for the spatial transform layer, all other settings and model architecture remain the same as our method. 

%It performs worse than our model in both up-sampling scale factors 4 and 8. In our experiment, we observe that the spatial transform layer tries to capture large area at the beginning, then focuses on a small scale at last few steps. This seems a reasonable strategy for image enhancement, but the spatial transform layer fails to capture the structure priors of the face images, possibly due to the large search space of the possible attention route, and thus leads to the inferior performance.

\subsection{Visualization of attended regions}
We visualize the detailed steps of how the agent works in our Attention-FH framework. We show the sequences of intermediate local enhancement results, as well as the attended regions located by the agent. As shown in Fig.~\ref{fig:attentive_example}, our Attention-FH is able to first locate the corner of the images, these regions are usually flat background and are easy to be enhanced without the knowledge of particular face characteristic. Secondly, the model turns to attend the facial components such as ear, eye, nose and mouth. Finally, the model refines detailed and high-frequency areas at last few steps.%, after enhancing facial components and other regions of the images. 

\subsection{Complexity of recurrent hallucination}
We compare the computation cost of our method with other one-pass full image SR methods. The testing time of a single $128\times128$ image running on TITANX is reported in table~\ref{table:time}. SRCNN\cite{dong2014learning} is a 3-layers CNN and it is the fastest among the compared methods. VDSR\cite{deepSR} use a very deep CNN with 20-layers, though it achieve state-of-art performance, it requires longer processing time. Our local enhancement net is an 8 layers CNN. Though it requires multiple pass, the running time is still comparable. Importantly, the extra recurrent iteration in our method is performed on patches, i.e. one forward pass per patch, it performs faster than other SR models\cite{deepSR}.

%More examples are shown in supplementary materials.

\section{Conclusion}
%In this paper, we propose a novel Attention-aware Face Hallucination (Attention-FH) framework and optimize it with deep reinforcement learning. In contrast to existing face hallucination methods, we cast the face hallucination problem as Markov decision process and categorize it into two tasks. This model is trained with delayed global reward to achieve top-down reasoning. Extensive experiments demonstrate that our model not only achieves state-of-the-art performance on popular evaluation datasets, but also has a better visual result. In future, we will try to extend our model to a more general form, which can handle other low-level vision processes.
In this paper, we propose a novel Attention-aware Face Hallucination (Attention-FH) framework and optimize it using deep reinforcement learning. In contrast to existing face hallucination methods, we explicitly incorporate the rich correlation cues among different facial parts by casting the face hallucination problem as a Markov decision process. 
%In each step, our framework perform two tasks: patch attention and enhancement. We train the policy network with deep reinforcement learning method that can incorporate delayed global reward to achieve top-down reasoning, and train the local enhancement network with local supervised loss to stabilize the reinforcement optimization. 
Extensive experiments demonstrate that our model not only achieves state-of-the-art performance on popular evaluation datasets, but also demonstrates better visual results. In future, we will try to extend our model to a more general form, which can handle other low-level vision problems.

\clearpage
{\small
\bibliographystyle{ieee}
\bibliography{egbib}
}

\end{document}